%% file: iclr2027_conference.tex
\documentclass{article} 
\usepackage{iclr2027_conference,times}

\input{math_commands.tex}

\usepackage{hyperref}
\usepackage{url}
\usepackage{graphicx}
\usepackage{booktabs}
\usepackage{multirow}
\usepackage{capt-of}
\usepackage{wrapfig}
\usepackage{algpseudocode}
\usepackage{xcolor}

\AddToHook{env/figure/begin}{\setlength{\abovecaptionskip}{8pt}}
\AddToHook{env/figure*/begin}{\setlength{\abovecaptionskip}{8pt}}

\title{\textsc{DeCoPrune}: Efficient KV-Cache Pruning for Autoregressive Video Diffusion via Denoising Consistency}

\author{%
\parbox{\dimexpr\textwidth-2\tabcolsep\relax}{\centering
\vspace{16pt}
\textbf{Zeqi Xiao}$^{1,*}$\quad
\textbf{Qingle Liu}$^{1,2,*}$\quad
\textbf{Kaiwen Zhang}$^{1}$\\[3pt]
\textbf{Yifan Zhou}$^{1}$\quad
\textbf{Zihan Ding}$^{3}$\quad
\textbf{Xingang Pan}$^{1}$\\[6pt]
{\normalfont $^{1}$S-Lab, Nanyang Technological University}\\[2pt]
{\normalfont $^{2}$Tsinghua University\quad $^{3}$Princeton University}\\[5pt]
{\normalfont\small $^{*}$Equal contribution.}%
}}

\hypersetup{
  colorlinks=true,
  linkcolor=blue,
  citecolor=blue,
  urlcolor=blue,
  pdftitle={DeCoPrune: Efficient KV-Cache Pruning for Autoregressive Video Diffusion via Denoising Consistency},
  pdfauthor={Zeqi Xiao, Qingle Liu, Kaiwen Zhang, Yifan Zhou, Zihan Ding, Xingang Pan}
}

\iclrfinalcopy
\begin{document}

\maketitle
\lhead{Preprint}
\vspace{-28pt}

\begin{center}
\centering
\includegraphics[width=\linewidth]{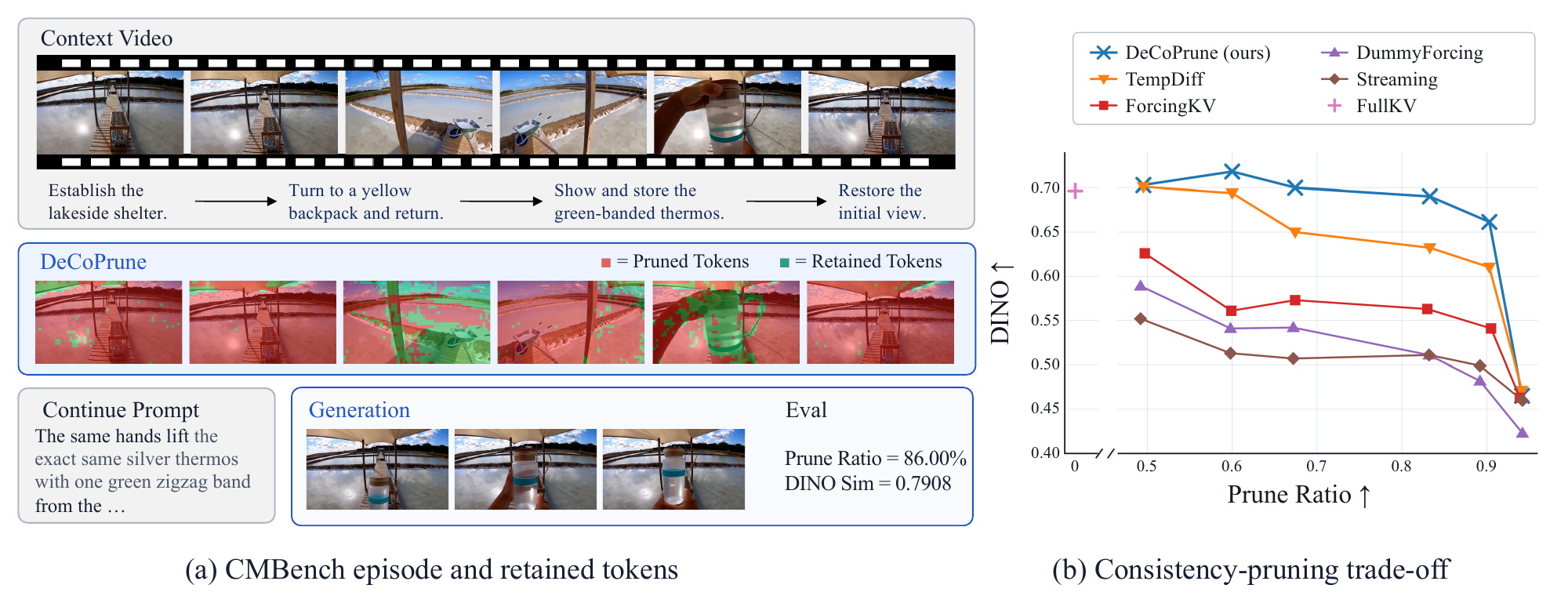}
\captionof{figure}{\textbf{Overview of \textsc{CMBench} and \textsc{DeCoPrune}.} (a) A representative \textsc{CMBench} episode showing which historical tokens \textsc{DeCoPrune} retains and reuses during continuation. (b) Consistency--pruning-ratio trade-off across pruning methods. At comparable pruning ratios, our method achieves higher DINO consistency than the evaluated compression baselines.}
\label{fig:teaser}
\end{center}

\input{sections/abstract}
\input{sections/introduction}
\input{sections/related_works}
\input{sections/methods}
\input{sections/experiments}
\input{sections/conclusion}

\newpage
\section*{AI Use Statement}
For manuscript preparation, we used generative AI tools only to aid writing
and polish language. All AI-assisted text was reviewed and verified by the
authors, who take full responsibility for the final content and claims of
this work. The use of H3 to construct synthetic benchmark videos is described
in the benchmark construction and Appendix~\ref{app:video_source}.

\section*{Ethics Statement}
This work studies efficient, context-consistent video generation for research
purposes. Such techniques may also lower the cost of producing misleading
synthetic videos. Consistency with a video context does not establish factual
accuracy or authenticity, and generated videos should be clearly identified
as synthetic. The use and distribution of real-world footage should respect
privacy, consent, and applicable licenses.

\section*{Reproducibility Statement}
Section~\ref{sec:decoprune} specifies the consistency score and cache-update
rule, and Section~\ref{sec:implementation_details} describes their application
to observed contexts and the supporting implementation choices. The
\textsc{CMBench} subsection details benchmark construction and evaluation,
while the experimental settings report the backbone, hardware, denoising
schedule, pruning threshold, and cache budgets.
Appendices~\ref{app:rope_reindexing} and~\ref{app:head_specialization}
provide the RoPE re-indexing and head-specialization details. The
supplementary material includes benchmark examples and qualitative video
comparisons to support inspection of the reported behavior.

\bibliography{iclr2027_conference}
\bibliographystyle{iclr2027_conference}

\clearpage
\appendix
\input{sections/appendix}

\end{document}

%% file: math_commands.tex
\usepackage{amsmath,amsfonts,bm}

\def\eqref#1{equation~\ref{#1}}

\def\1{\bm{1}}

\DeclareMathAlphabet{\mathsfit}{\encodingdefault}{\sfdefault}{m}{sl}
\SetMathAlphabet{\mathsfit}{bold}{\encodingdefault}{\sfdefault}{bx}{n}



%% file: sections/abstract.tex
\begin{abstract}
Autoregressive video diffusion naturally supports streaming generation and
interactive control, but its KV cache grows continuously with the generated
history. Existing compression strategies either discard history using fixed
windows or select tokens through local attention and similarity signals,
which do not directly measure whether the current chunk contributes information
beyond the retained context. We introduce \textsc{DeCoPrune}, a
training-free method that treats cache compression as a denoising-consistency
problem. We find empirically that denoising difficulty provides a useful proxy
for a token's value in long-term retention: tokens with larger step-to-final
discrepancies tend to carry visual evidence that is less predictable from
the retained context.
Based on this finding, \textsc{DeCoPrune} measures each current-chunk token's
denoising difficulty using the discrepancy between its intermediate clean
prediction and final denoised value, retaining high-discrepancy tokens in the
long-term cache while pruning those with low discrepancy. To evaluate
information retention, we introduce \textsc{CMBench}, comprising 58 approximately
one-minute generated or real-world context episodes and 116
\textit{Reappear} or \textit{Revisit} continuation tasks that require recalling
specific previously observed objects or scenes.
Experiments with LingBot World v2 show that \textsc{DeCoPrune} preserves near-FullKV long-range recall while pruning over 85\% of historical KV tokens and accelerating continuation generation by over 4$\times$, substantially outperforming the evaluated compression baselines at comparable budgets.
These results indicate that denoising consistency can serve as a
model-intrinsic signal for retaining
long-range information while reducing autoregressive inference
cost.
{\urlstyle{tt}
Our project homepage is
\raisebox{-0.15em}{\includegraphics[height=1em]{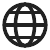}}\,\url{https://decoprune.github.io}.
The code is available at
\raisebox{-0.15em}{\includegraphics[height=1em]{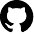}}\,\url{https://github.com/DeCoPrune/CMBench},
and the benchmark at
\raisebox{-0.15em}{\includegraphics[height=1em]{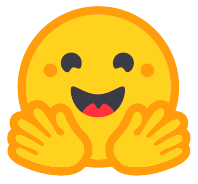}}\,\url{https://huggingface.co/datasets/Aoraku/CMBench}.}
\end{abstract}


%% file: sections/introduction.tex
\section{Introduction}

Recent video generators have achieved high visual fidelity and motion
realism~\citep{wan2025}. Autoregressive (AR) video diffusion generates videos
sequentially, often in temporal chunks, naturally supporting streaming generation and interactive
control~\citep{selfforcing2025,lingbotworld2_2026}.
In the chunk-based AR models, each chunk attends to cached KVs from the preceding video history, so memory
and attention costs grow steadily with the rollout. Retaining the full history
eventually becomes impractical, while aggressive truncation removes visual
evidence needed for long-range consistency.

The challenge is therefore to remove redundant history while preserving
evidence needed for long-range consistency. Some approaches simply discard
most older tokens~\citep{streamingvlm2026}, risking substantial information
loss. Temporal-difference selection instead removes tokens based on similarity
between neighboring frames~\citep{hwang2024everest,fu2025framefusion}, but such
local comparisons cannot identify repetition across distant parts of the
history. Redundancy is not confined to adjacent frames: objects and scenes
can recur throughout a long context, even after substantial intervening
changes. We therefore need a retention criterion that evaluates the current
chunk against the entire retained context, distinguishing evidence already
represented in the cache from content that warrants additional storage.

Our starting observation is that informative prior context reduces the discrepancy
between intermediate clean predictions and final denoised outputs across
four backbones (Figure~\ref{fig:motivation}). This suggests a simple pruning
heuristic: if a token's intermediate prediction is already close to its final
output, the retained context may already provide much of the information
needed to predict it, making its KV entry a candidate for removal. We therefore
prioritize retaining tokens with larger discrepancies and prune those with
smaller discrepancies.

Based on this observation, we introduce \textsc{DeCoPrune}, a training-free
online KV-cache compression method (Figure~\ref{fig:overview}). For each chunk,
we compare an intermediate clean prediction with the final output from the
same denoising trajectory, then threshold the token-wise mean squared
discrepancy into a retention mask. A shared mask selects high-discrepancy KV
entries across layers, compacting the KV cache and shortening the history attended to by
future chunks rather than merely masking attention weights. Online scoring
reuses predictions from the normal denoising trajectory, requiring no extra model
evaluations or training. For observed video contexts without a denoising
trajectory, we re-noise each chunk and compare its context-conditioned clean
prediction with the observed chunk.

Existing benchmarks assess general video quality~\citep{vbench2024} or
consistency with historical context~\citep{membench2026,memobench2026}. However, they do not explicitly evaluate targeted recall of specific previously observed objects or scenes, and visual quality can
remain stable as historical recall deteriorates
(Figure~\ref{fig:probe_threshold_selection}(b)). We therefore introduce Context Memory
Benchmark (\textsc{CMBench}), comprising 58 approximately one-minute generated
or real-world contexts and 116 \textit{Reappear}/\textit{Revisit} continuation
tasks. These tasks require the model to recover specific previously observed visual content, enabling reference-grounded evaluation of long-range recall under controlled historical-KV compression.

\textsc{DeCoPrune} achieves 0.6701 DINO on LingBot World
v2 with an 85.43\% reduction in cumulative
historical KV token counts and a 4.14$\times$ continuation-generation speedup
over FullKV.
Its head-specialized variant reaches 0.6783 DINO at an 86.19\% pruning ratio,
approaching the 0.6803 score of \textit{FullKV} and exceeding the evaluated
compression baselines at comparable budgets.

In summary, our contributions are threefold:
\begin{itemize}

\item \textsc{DeCoPrune}: a training-free KV-cache pruning method that uses context-conditioned step-to-final denoising discrepancy as a model-intrinsic signal for retaining informative historical tokens;

\item \textsc{CMBench}: a reference-grounded benchmark with Reappear and Revisit tasks for evaluating long-range visual recall under controlled historical-KV compression;

\item Extensive experiments demonstrating near-FullKV long-range recall with over 85\% historical-KV pruning and more than 4$\times$ continuation-generation speedup, while substantially outperforming the evaluated compression baselines at comparable budgets.

\end{itemize}

%% file: sections/related_works.tex
\section{Related Work}

\paragraph{Autoregressive Long-Video Generation.}
Video diffusion models
~\citep{dit2023,opensoraPlan2024,hunyuanvideo2024,wan2025,magi12025,skyreelsv2_2025,longcatvideo2025}
extend to long rollouts through history conditioning~\citep{diffusionforcing2024,historyguided2025},
distillation and self-rollout training~\citep{causvid2025,selfforcing2025,causalforcing2026},
and long-horizon training and sampling
~\citep{rewardforcing2025,infinityrope2025,selfforcingpp2025,causalforcingpp2026,stablevideoinfinity2025,rollingforcing2025,ringforcing2026}.
Long rollouts also benefit from recurrent or compressed states
~\citep{malt2025,framepack2025,videossm2025,helios2026},
while action conditioning enables interactive worlds
~\citep{diamond2024,gamengen2024,mineworld2025,matrixgame2_2025,lingbotworld2_2026}.
Feature/cache reuse~\citep{pab2025,teacache2025,toca2025,duca2024,flowcache2026,worldcache2026,ca2vdm2025}
and sparse attention~\citep{sparsevideogen2025,lightforcing2026,sparseforcing2026}
reduce denoiser computation; we instead study historical visual evidence
retained in a frozen generator's native KV cache.

\paragraph{KV-Cache Compression and Video Memory.}
Cache compression uses windows, importance scores, and head/layer budgets
~\citep{streamingllm2024,h2o2023,snapkv2024,pyramidkv2024,duoattention2024}.
Video methods extend window-based compression
~\citep{longlive2025,deepforcing2025,fademem2026,packforcing2026},
head specialization~\citep{dummyforcing2026,forcingkv2026,headforcing2026,pyramidforcing2026},
content-aware selection~\citep{tempcache2026,packcache2026,futureforcing2026,focusedforcing2026,pafukv2026,densitykv2026},
and low-rank or quantized representations~\citep{videomla2026,quantvideogen2026}.
Related approaches use key similarity or motion novelty~\citep{yi2026worldkv,recapforcing2026}.
Explicit memories rely on retrieval
~\citep{xiao2026worldmem,yu2025context,li2025vmem,wu2026video,vrag2025,longliverag2026},
learned context
~\citep{malt2025,memlearner2026,tinyhistory2026,memorizegenerate2025,streamingt2v2025,memorypack2025},
or entity-centric and refinement mechanisms
~\citep{zhang2025storymem,videomemory2026,echoforcing2026,slotmemory2026,a2rd2026}.
Positional re-indexing follows~\citet{wu2026addressable}.
\textsc{DeCoPrune} instead scores step-to-final denoising change, without
an auxiliary salience model, retrieval index, or motion estimator.

\paragraph{Context-Memory Evaluation.}
Benchmarks assess general video quality and plausibility
~\citep{vbench2024,vbenchlong2025,storyeval2024,narrlv2025,worldmodelbench2025},
historical consistency
~\citep{memorizegenerate2025,membench2026,videomemory2026,a2rd2026},
post-occlusion recovery~\citep{memobench2026},
and scene revisitation or action control~\citep{wu2026addressable,mind2026,r2mbench2026}.
\textsc{CMBench} couples reference-grounded \textit{Reappear}/\textit{Revisit}
recall to controlled historical-KV compression in minute-scale real and
generated contexts.

%% file: sections/methods.tex
\section{Methodology}

\subsection{Denoising Consistency for KV-Cache Pruning}
\label{sec:decoprune}

\begin{figure}[t]
\centering
\includegraphics[width=\linewidth]{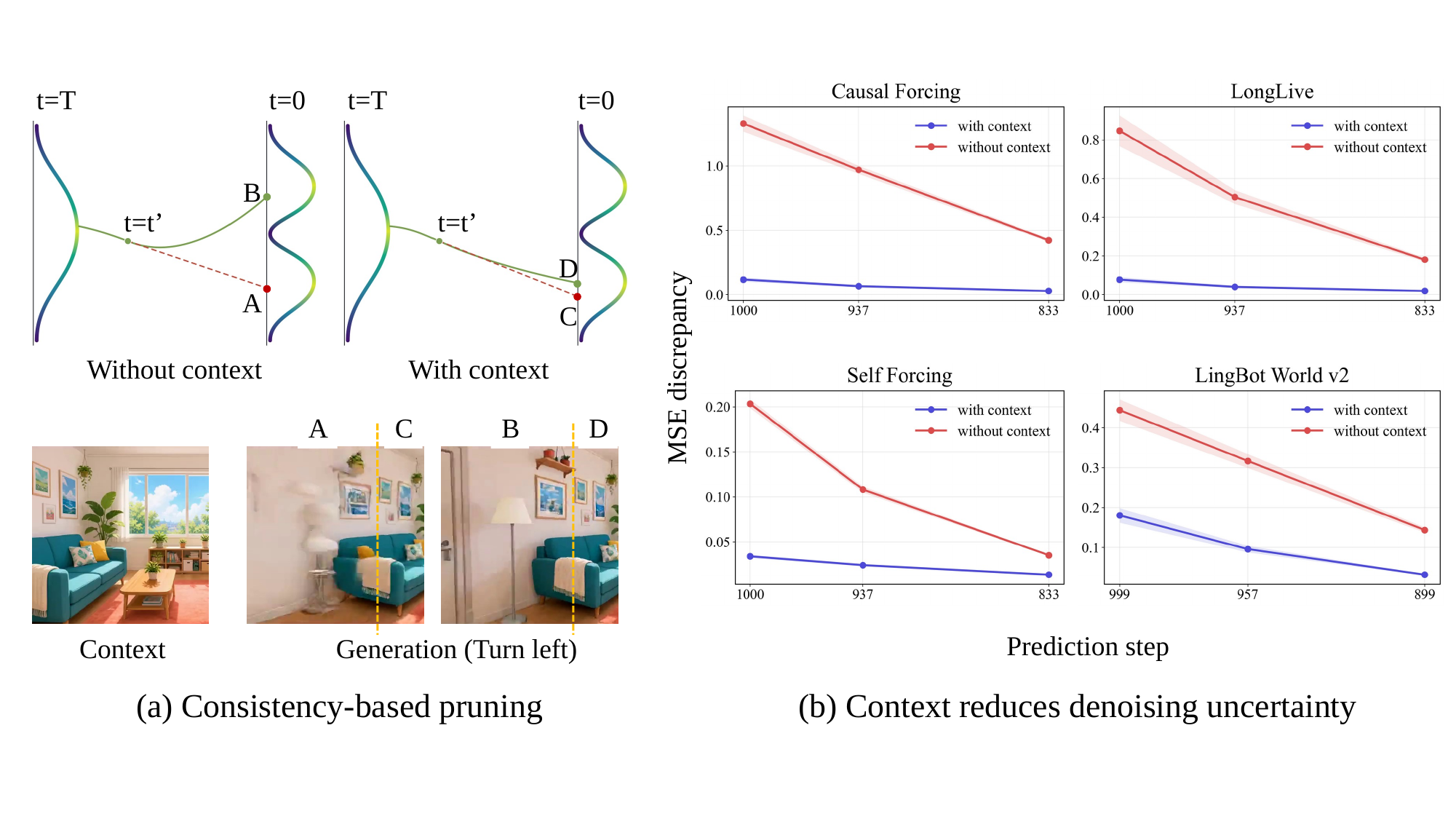}
\vspace{-24pt}
\caption{\textbf{Denoising consistency and context redundancy.}
\textbf{(a)} At an intermediate denoising state $t=t'$, a local step-to-final
extrapolation (red dashed) can deviate from the final denoised
state without context ($A\!\to\!B$). Relevant context concentrates the
conditional outcome, aligning the extrapolation more closely with the final
state ($C\!\to\!D$). Our policy treats content that is consistently predictable
from existing context as redundant, pruning low-discrepancy tokens while
retaining high-discrepancy tokens. \textbf{(b)} Curves report the
MSE between the $x_0$ prediction at each intermediate denoising step and the
final $x_0$. The Causal Forcing, LongLive, and Self-Forcing plots compare runs
with and without the available context KV: first-frame KV for Causal Forcing
and Self-Forcing, and preceding-chunk KV for LongLive. For LingBot World v2,
which does not support text-to-video generation, both runs use an initial frame
and rotate the camera: ``with context'' means that the content revealed after
rotation appeared earlier in the context, whereas ``without context'' means
that it did not. In all four comparisons, the with-context condition has lower
error across the plotted steps, consistent with our criterion.}
\label{fig:motivation}
\end{figure}

\paragraph{Setup and hypothesis.}
An autoregressive video diffusion model generates latent chunks
$\mathbf{x}_i\in\mathbb{R}^{N\times d}$ conditioned on input $\mathbf{c}_i$
and the preceding cache
$\mathcal{H}_{<i}=\{(\mathbf{K}_{\ell,<i},\mathbf{V}_{\ell,<i})\}_{\ell=1}^{L}$,
where $N$ is the number of tokens, $d$ their dimension, and $L$ the number of
layers. Our hypothesis is that tokens already predictable from this cache
tend to reach stable clean predictions earlier, whereas tokens carrying
information not explained by the retained context tend to require greater refinement.
When the retained context already determines a token's content, denoising
primarily recovers information available in the conditioning cache. In
contrast, content not determined by the context may undergo greater refinement
along the trajectory. Figure~\ref{fig:motivation} provides supporting
context-conditioned measurements across four backbones. We therefore use
step-to-final prediction change as a model-intrinsic proxy for contextual
redundancy, without training an importance predictor or computing external
retrieval embeddings.

\paragraph{Token scoring and selection.}
During the normal generation of chunk $i$, we record the clean prediction
$\hat{\mathbf{x}}_{0,i}(\tau^*)=
D_\theta(\mathbf{z}_i^{\tau^*},\tau^*;\mathbf{c}_i,\mathcal{H}_{<i})$
at probe timestep $\tau^*$, where $\mathbf{z}_i^{\tau^*}$ is the noisy latent
on that trajectory. Once the same trajectory produces
$\mathbf{x}_i^{\mathrm{final}}$, we compute
\begin{equation}
\ell_{i,p}=\frac{1}{d}\left\|
\hat{\mathbf{x}}_{0,i,p}(\tau^*)-\mathbf{x}_{i,p}^{\mathrm{final}}
\right\|_2^2,
\qquad
m_{i,p}=\mathbb{I}[\ell_{i,p}>\gamma],
\quad p=1,\ldots,N.
\label{eq:token_score}
\end{equation}
Here $m_{i,p}=1$ means retention: high-discrepancy tokens enter the long-term
cache, while low-discrepancy tokens are treated as redundant. The threshold
$\gamma$ controls the retention--compression trade-off.

\begin{figure}[t]
\centering
\includegraphics[width=\linewidth]{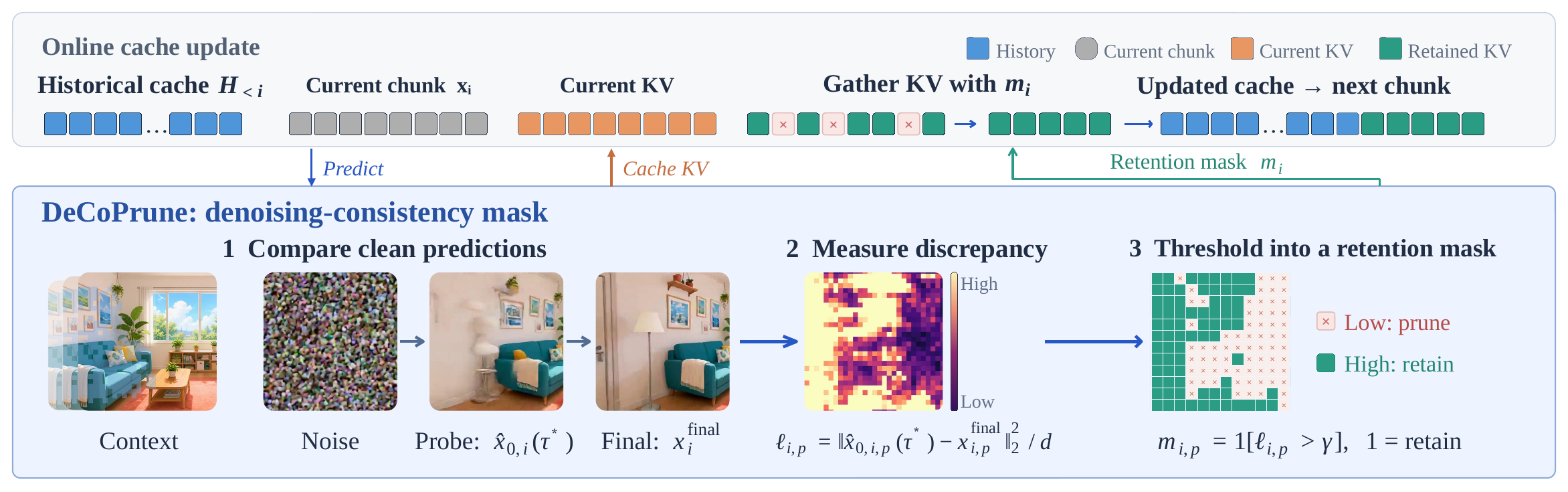}
\caption{\textbf{\textsc{DeCoPrune} overview.} A probe and final prediction
from the same denoising trajectory yield a token-retention mask. The mask
physically compacts historical KVs after the recent-window delay.}
\label{fig:overview}
\end{figure}

\paragraph{Physical cache update.}
As Figure~\ref{fig:overview} shows, we protect the initial sink chunk and keep
the most recent $w$ chunks' KVs dense, storing their masks for later use.
When a non-sink chunk $i$ leaves this window, we gather the selected entries
in every layer:
\begin{equation}
(\widetilde{\mathbf{K}}_{\ell,i},\widetilde{\mathbf{V}}_{\ell,i})
=(\mathbf{K}_{\ell,i}[\mathbf{m}_i],\mathbf{V}_{\ell,i}[\mathbf{m}_i]),
\qquad \ell=1,\ldots,L.
\label{eq:kv_compaction}
\end{equation}
The compacted entries replace that chunk's dense record in the cache used by
subsequent generation. The shared mask shortens the physical KV sequence
rather than only masking attention weights. This keeps the full local evidence
available to immediate successors, while informative tokens from earlier
chunks remain accessible through the compressed cache. The generator remains
frozen, and online scoring reuses predictions from its normal denoising
trajectory.

\subsection{CMBench}

\begin{figure}[t]
\centering
\includegraphics[width=\linewidth,trim=36bp 0bp 36bp 58bp,clip]{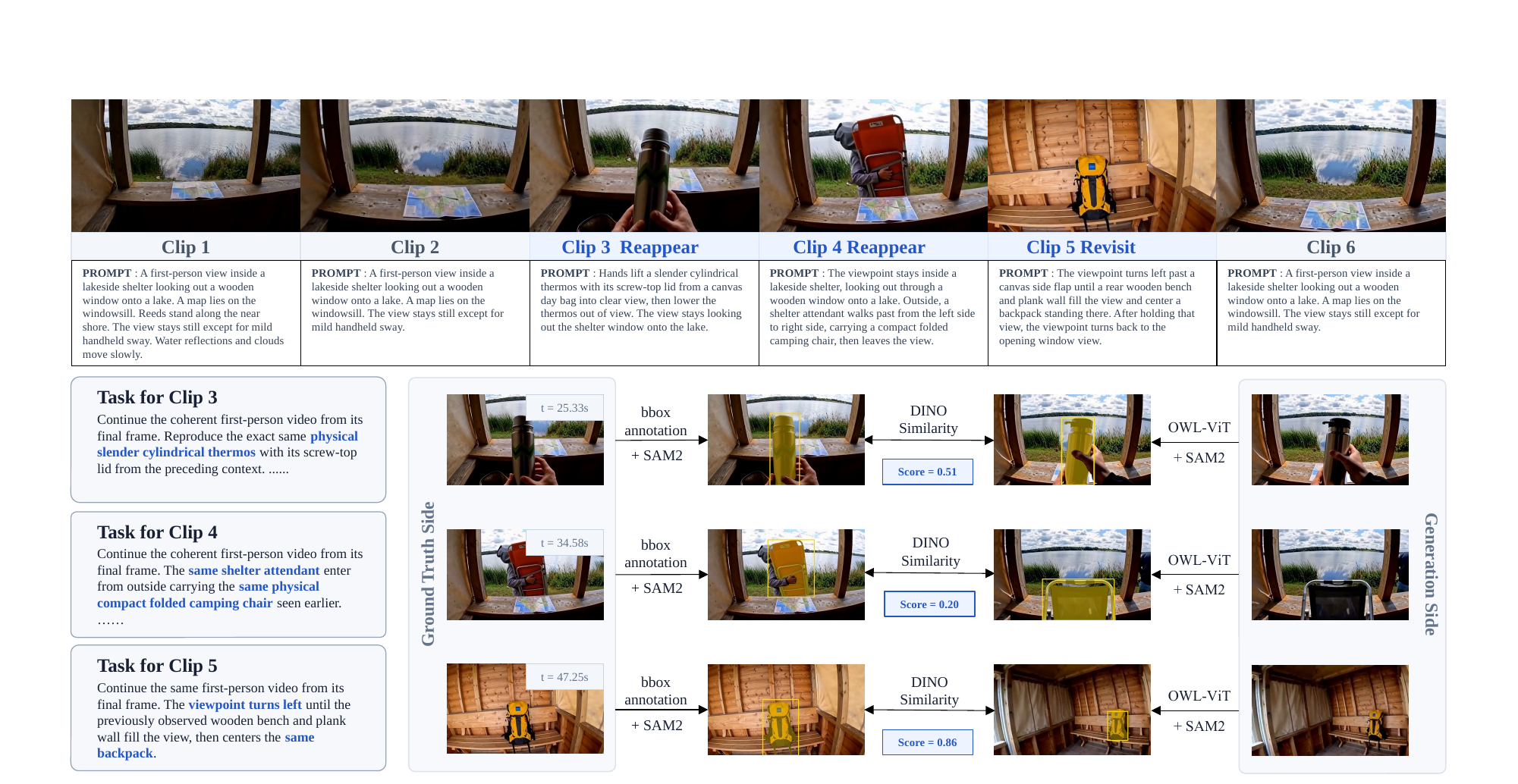}
\caption{\textbf{Overview of \textsc{CMBench}.} \textbf{Top:} a one-minute
generated context assembled from six prompted clips; three target events yield
three continuation tasks. \textbf{Bottom:} the corresponding continuation
prompts and evaluation pipeline. The reference target and its generated
counterpart are localized and segmented, then compared using DINO similarity.}
\label{fig:cmbench_overview}
\end{figure}

\paragraph{Benchmark setting.}
\textsc{CMBench} evaluates recall of previously observed visual evidence under controlled historical-KV compression. It comprises 116 continuation tasks across 58
episodes: 50 H3-generated episodes provide 102 tasks, and eight real-world
episodes provide the remaining 14. Each episode is built around an
approximately one-minute video context and contains
multiple target events; pairing one event with its continuation prompt defines
a task. As illustrated in Figure~\ref{fig:cmbench_overview}, each prompt asks
the generator to reproduce an object, person, or view that appeared in the
context, directly testing whether that visual evidence is retained and can be
used during continuation.

The generated contexts are constructed with MiniMax H3~\citep{minimaxh3_2026}
to control event content and timing, while real-world contexts provide
complementary natural footage under the same task and evaluation protocol. A
generated context concatenates six 10-second clips produced with first--last
frame conditioning; the boundary frame is propagated between adjacent clips
to maintain continuity. Each clip contains at most one complete target event,
preventing independently generated boundaries from altering the event. Multiple
events within a context act as distractors for one another. Appendix~\ref{app:video_source}
reports the source-stratified analysis.

\paragraph{Continuation tasks.}
\textit{Reappear} asks a person or object observed in the context to appear
again. Success requires reproducing the same entity and appearance, rather than
a plausible instance of the same category. \textit{Revisit} first shows a
camera transition from scene $A$ to scene $B$ and back to $A$, then asks the
continuation to revisit $B$. Scene $B$ contains a salient object that serves as
an evaluation anchor. In generated contexts, the complete $A\!\rightarrow\!B
\!\rightarrow\!A$ event lies within one 10-second clip so that $B$ remains a
consistent reference. Both tasks therefore reduce to recalling a target that
was visually observed in the context.

\paragraph{Evaluation protocol.}
For each task, we annotate a reference frame and target bounding box in the
context. As Figure~\ref{fig:cmbench_overview} shows, OWL-ViT~\citep{owlvit}
localizes the target in each generated frame and SAM 2~\citep{sam2} segments the
reference and generated targets. Following subject-fidelity evaluation in
personalized generation~\citep{dreambooth2023}, we compare DINOv2
embeddings~\citep{dinov2} of the resulting crops. For reference crop $r$ and
generated crop $g_t$ at continuation frame $t$, the task score is
\begin{equation}
S_{\mathrm{DINO}}
=
\max_t\operatorname{cos}\!\left(
f_{\mathrm{DINO}}(r), f_{\mathrm{DINO}}(g_t)
\right).
\label{eq:cmbench_dino_score}
\end{equation}
The maximum allows the requested event to occur at any point in the
continuation; the score is zero if the target is never detected. For
\textit{Revisit}, the salient object in scene $B$ is the target, enabling the
same protocol for both task types. We report the mean score over all tasks.

We also measure the effective sequence-level pruning ratio (PR) during
autoregressive generation. Let $k_{i,\ell,h}$ be the number of historical KV
token positions physically visible to attention head $h$ in layer $\ell$ when
denoising generated chunk $i$, and let $k_{i,\ell,h}^{\mathrm{full}}$ be the
corresponding number under FullKV. For a continuation of $T$ generated chunks
in a model with $L$ layers and $H$ attention heads, we define
\begin{equation}
\mathrm{PR}
=
1-
\frac{\sum_{i=1}^{T}\sum_{\ell=1}^{L}\sum_{h=1}^{H}k_{i,\ell,h}}
{\sum_{i=1}^{T}\sum_{\ell=1}^{L}\sum_{h=1}^{H}k_{i,\ell,h}^{\mathrm{full}}}.
\label{eq:cmbench_pruning_ratio}
\end{equation}
PR aggregates historical KV counts across chunks, layers, and heads,
excluding the current noisy chunk from both sums. We report its arithmetic
mean over cases; it measures cumulative historical-token reduction, not peak
memory savings or total computation. FPS and speedup measure continuation
generation only, excluding prefix processing. Appendix~\ref{app:pr_aggregation}
details the aggregation. We also report standard VBench quality
metrics~\citep{vbench2024}.

\subsection{Implementation Details}
\label{sec:implementation_details}

We complement the core pruning criterion with prefix pruning, RoPE
re-indexing, a recent local window, and an optional head-specialized variant,
which respectively support observed contexts, long-range addressability,
short-term continuity, and higher compression.

\paragraph{Prefix pruning.}
For an already observed video context, the original denoising trajectory is
not available at compression time. We therefore re-noise each finalized
context chunk to $\tau^*$, evaluate its $\mathbf{x}_0$ prediction at its
original absolute temporal position conditioned on the preceding retained
cache, and apply Equation~\ref{eq:token_score}, using the observed clean chunk as the reference. Processing the
context chunks sequentially constructs a compressed KV cache for subsequent
autoregressive continuation.

\paragraph{RoPE re-indexing.}
Large query--key temporal offsets can make retained history difficult to
retrieve. Following \citet{wu2026addressable}, we compress the
temporal RoPE coordinates of older context keys into a fixed virtual span
while leaving the most recent frames at their original positions. Conceptually,
for a context frame at position $t$ and continuation start $q_0$, we use
$\widetilde t(t)=q_0-g(q_0-t)$, where $g$ preserves recent offsets and linearly
compresses older ones. This preserves temporal order and changes only key
phases, without modifying the selected tokens, values, or spatial RoPE
coordinates. The exact mapping and phase correction are given in
Appendix~\ref{app:rope_reindexing}.

\paragraph{Head-specialized variant.}
As an extension orthogonal to our token-selection criterion,
\textsc{DeCoPrune--HS} adopts the attention-based head partition of
\textit{ForcingKV}~\citep{forcingkv2026} to increase the pruning ratio. Dynamic
heads apply \textsc{DeCoPrune}, whereas static heads use \textit{Streaming}.
This head-wise specialization improves compression without altering the core
denoising-consistency criterion. The offline head classification, per-group
cache policies, and layer-0 treatment are detailed in
Appendix~\ref{app:head_specialization}.

%% file: sections/experiments.tex
\section{Experiments}

\subsection{Settings}

\paragraph{Backbone and implementation.}
We use LingBot World v2~\citep{lingbotworld2_2026} as the primary backbone for our experiments. All experiments are conducted on four NVIDIA H200 GPUs. For completeness, we also evaluate Self-Forcing, Causal Forcing, and LongLive~\citep{selfforcing2025,causalforcing2026,longlive2025}; these backbones are not well suited to minute-scale long-context generation. Appendix~\ref{app:backbone_selection} compares their FullKV performance with minute-scale contexts. To additionally test our pruning method on another backbone, Appendix~\ref{app:longlive_short_context} (Table~\ref{tab:longlive_short_context}) evaluates LongLive on 18 continuation cases with 10-second contexts, where \textsc{DeCoPrune} outperforms the other pruning baselines in both DINO and PR.

\paragraph{Baselines and evaluation.}
We compare against \textit{FullKV}, which retains the complete attention context, and a \textit{Streaming} baseline following StreamingVLM and LongLive~\citep{streamingvlm2026,longlive2025}, which keeps only sink and recent tokens. \textit{DummyForcing}~\citep{dummyforcing2026} and \textit{ForcingKV}~\citep{forcingkv2026} apply different cache policies to different attention heads. \textit{TempDiff} is a temporal-difference baseline inspired by temporal-redundancy-aware token reduction~\citep{hwang2024everest,fu2025framefusion}: it compares each patch with its counterpart in the preceding frame and inserts the $K$ least similar patches into the KV cache. \textit{Random} preserves the same sink and recent windows as our method while sampling the remaining historical tokens at random. We evaluate both \textsc{DeCoPrune} and its head-specialized variant, \textsc{DeCoPrune--HS}, described in Section~\ref{sec:implementation_details}. \textsc{CMBench} is our primary evaluation, with the DINO score (reported on a 0--1 scale) measuring consistency level and PR and FPS measuring efficiency; we additionally report \textsc{VBench}~\citep{vbench2024} to assess general video quality.

\paragraph{Denoising schedule and probe.}
We use a four-step noise schedule $(999,957,899,702)$ derived from
\texttt{FlowUniPC} with a shift of $5$. The zero-based denoising-step indices
$1$, $2$, and $3$ correspond to noise timesteps $957$, $899$, and $702$,
respectively. We use index $2$ ($\tau^*=899$) as the probe step and
$\gamma=0.10$ as the pruning threshold. The default video configuration is
$832\times480$ pixels at 16 FPS, with four latent frames per chunk.

\subsection{Main Results}

\paragraph{Comparison protocol.}
All compression methods share the same sink and recent windows
(Appendix~\ref{app:head_specialization}).
\textit{Streaming} and \textit{DummyForcing} discard intermediate history;
we tune history-aware methods to comparable PR values. In our reproductions,
\textit{DummyForcing} uses cache
sizes 4/8/4 for its first/middle/last head groups, \textit{ForcingKV} retrieves
$K=256$ patches from the preceding 32 chunks, \textit{TempDiff} uses 1,657
cache entries, and \textit{Random} matches the PR of \textsc{DeCoPrune}. All
methods use the same RoPE re-indexing in the main comparison; Table~\ref{tab:reindex_ablation}
isolates its effect. Table~\ref{tab:cmbench_results} summarizes the results.

\begin{table}[t]
\centering
\caption{Main comparison on \textsc{CMBench} and \textsc{VBench}. DINO is reported on a 0--1 scale; results are averaged over three random seeds. FPS and speedup over FullKV measure continuation generation only, excluding prefix processing.}
\label{tab:cmbench_results}
\scriptsize
\setlength{\tabcolsep}{2.6pt}
\renewcommand{\arraystretch}{1.12}
\resizebox{\linewidth}{!}{%
\begin{tabular}{@{}lrrrrrrrr@{}}
\toprule
\multirow{2}{*}{Method} &
\multirow{2}{*}{DINO $\uparrow$} &
\multirow{2}{*}{PR $\uparrow$} &
\multirow{2}{*}{FPS $\uparrow$} &
\multirow{2}{*}{Speedup $\uparrow$} &
Temporal & Motion & Aesthetic & Image \\
& & & & &
Flickering $\uparrow$ & Smoothness $\uparrow$ &
Quality $\uparrow$ & Quality $\uparrow$ \\
\midrule
FullKV & 0.6803 & 0.00\% & 1.568 & 1.00$\times$ &
0.9499 & 0.9705 & 0.4633 & 0.7125 \\
Streaming~\citep{streamingvlm2026,longlive2025} & 0.4592 & 94.37\% & 9.591 & 6.12$\times$ &
0.9474 & 0.9698 & 0.4769 & 0.7103 \\
DummyForcing~\citep{dummyforcing2026} & 0.4461 & 98.03\% & 7.393 & 4.72$\times$ &
0.9425 & 0.9687 & 0.4677 & 0.6933 \\
ForcingKV~\citep{forcingkv2026} & 0.5313 & 80.62\% & 5.288 & 3.37$\times$ &
0.9437 & 0.9661 & 0.4610 & 0.7018 \\
TempDiff~\citep{hwang2024everest,fu2025framefusion} & 0.6229 & 86.36\% & 6.443 & 4.11$\times$ &
0.9425 & 0.9634 & 0.4677 & 0.7161 \\
\midrule
DeCoPrune (ours) & 0.6701 & 85.43\% & 6.489 & 4.14$\times$ &
0.9424 & 0.9650 & 0.4736 & 0.7067 \\
DeCoPrune--HS (ours) & 0.6783 & 86.19\% & 6.414 & 4.09$\times$ &
0.9428 & 0.9652 & 0.4634 & 0.7019 \\
\bottomrule
\end{tabular}%
}
\end{table}

\begin{figure}[t]
\centering
\includegraphics[width=\linewidth]{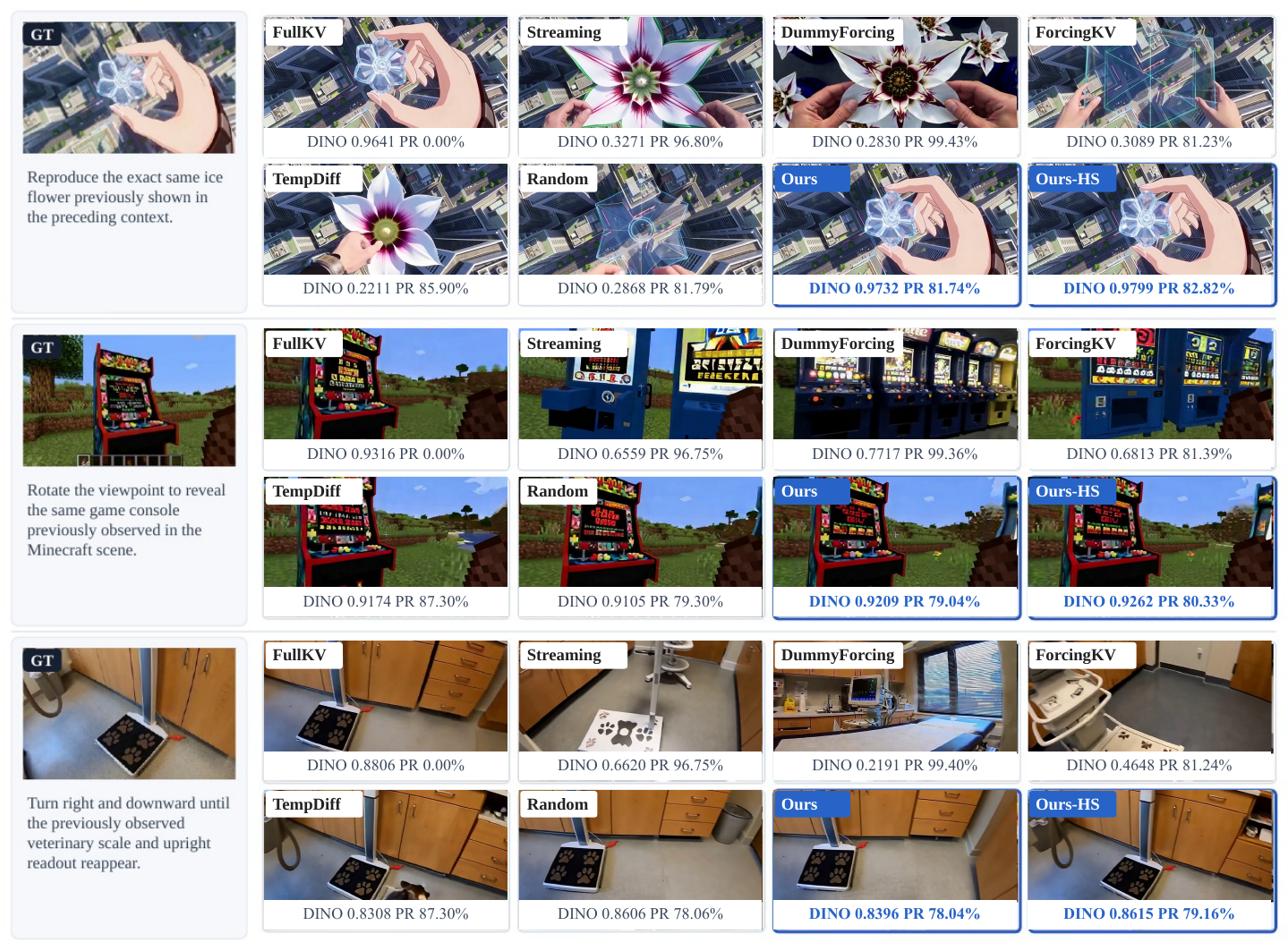}
\vspace{-4pt}
\caption{\textbf{Qualitative comparison under the matched settings used in Tables~\ref{tab:cmbench_results} and~\ref{tab:selection_direction_ablation}.}
We highly recommend viewing the video comparisons on the supplementary webpage
to better appreciate temporal consistency and visual detail.}
\label{fig:qualitative_comparison}
\end{figure}

\paragraph{\textsc{CMBench} results.}
Table~\ref{tab:cmbench_results} shows that DeCoPrune preserves near-FullKV
context consistency with an 85.43\% pruning ratio and a $4.14\times$
continuation-generation speedup over FullKV. FullKV remains the uncompressed reference with the
highest DINO in the main comparison. At similar PR and throughput, DeCoPrune
exceeds TempDiff by 0.0472 DINO, indicating that denoising-based selection
preserves useful evidence beyond a temporal-change heuristic.
DeCoPrune--HS further approaches FullKV with a 0.0020 DINO gap.
Streaming and DummyForcing are faster but sacrifice substantially more recall
by discarding intermediate history. These policies represent a different
compression--consistency trade-off, rather than matched-budget alternatives.
The matched-budget Random control is analyzed separately in
Table~\ref{tab:selection_direction_ablation}.
Figure~\ref{fig:qualitative_comparison} provides case-level comparisons under
the settings used in these tables.
The four \textsc{VBench} dimensions remain broadly comparable across policies,
providing a complementary assessment of general video quality.

\subsection{Ablation Study}

We ablate the three design choices underlying our context-management
pipeline: the timestep and threshold used to probe denoising consistency, the
positional re-indexing of retained history, and the direction of token
selection. Together, these studies examine whether the observed gains arise
from the proposed consistency signal rather than solely from the cache budget
or positional correction.

\begin{figure}[t]
\centering
\begin{minipage}[t]{0.55\linewidth}
\centering
\includegraphics[width=\linewidth]{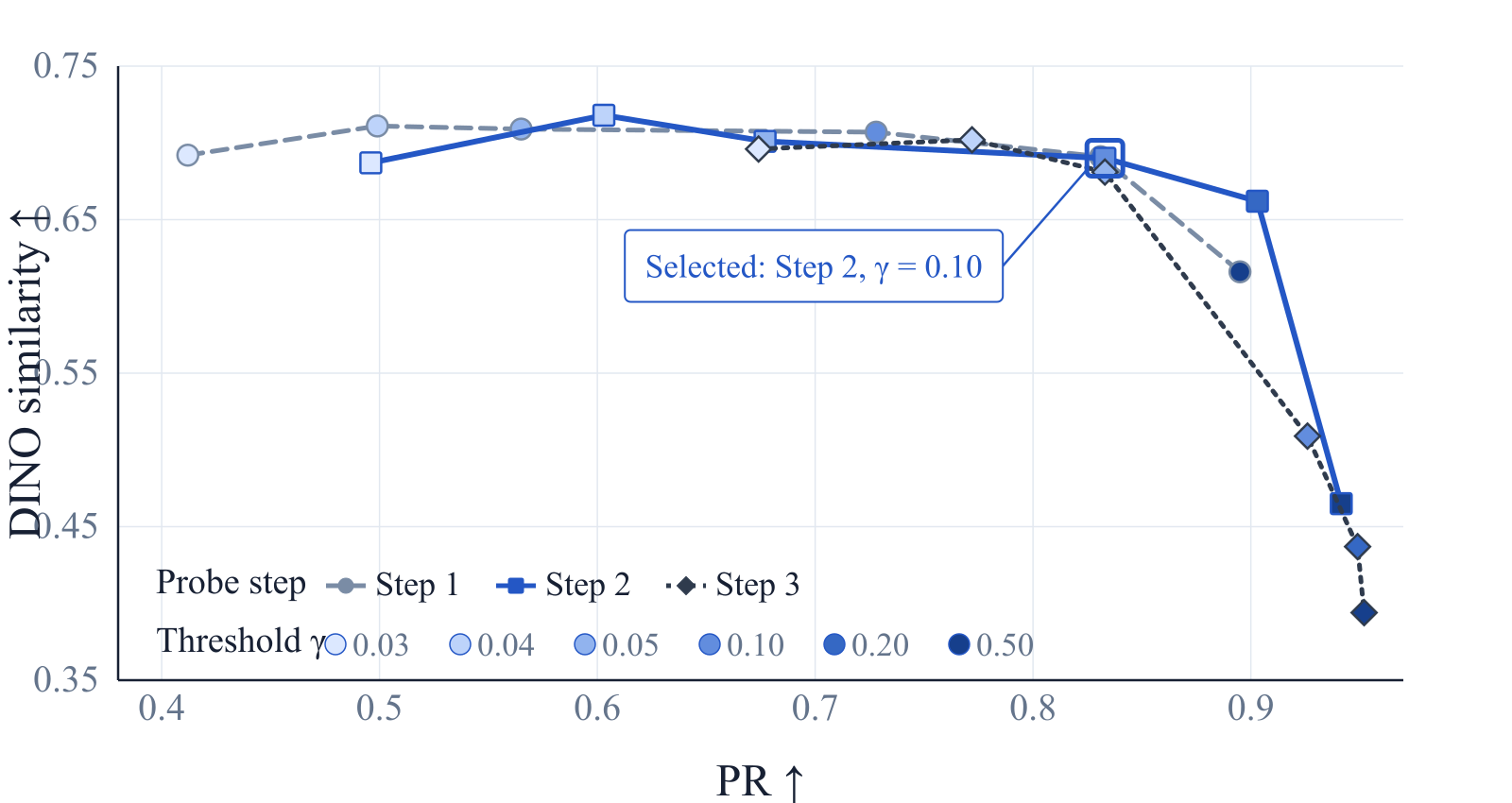}
\vspace{0pt}
\small (a) Probe timestep and pruning threshold
\end{minipage}\hfill
\begin{minipage}[t]{0.43\linewidth}
\centering
\includegraphics[width=\linewidth]{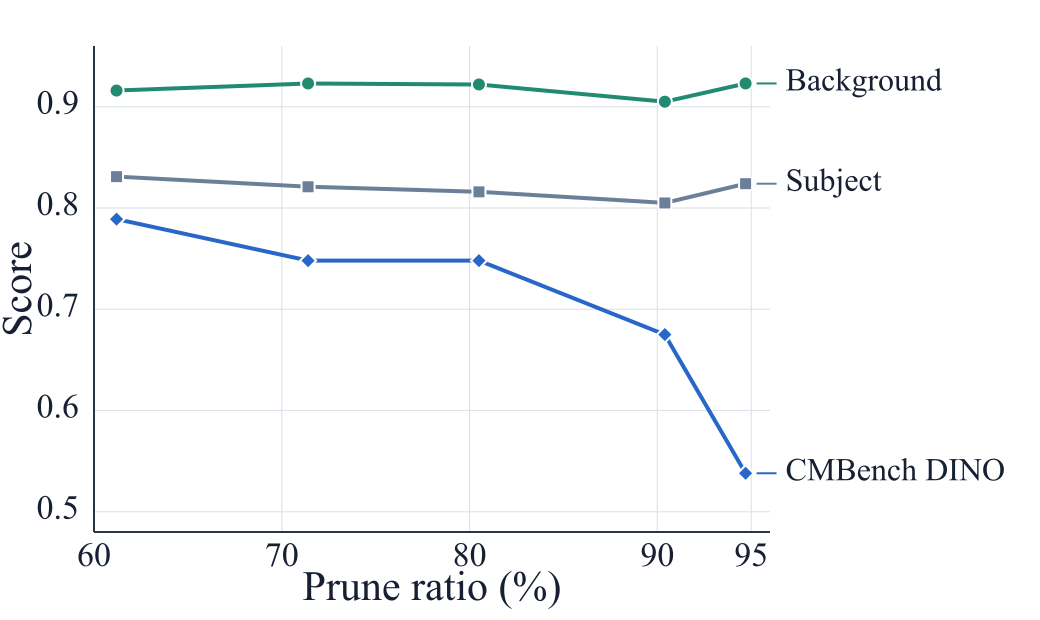}
\vspace{0pt}
\small (b) Sensitivity of consistency metrics
\end{minipage}
\caption{\textbf{Ablation and metric analysis on 13 independent cases.}
\textbf{(a)} Threshold $\gamma$ is swept from light to dark at each probe-step
index; the marked operating point is $s^*=2$ ($\tau^*=899$), $\gamma=0.10$.
\textbf{(b)} VBench subject/background consistency and CMBench DINO versus PR.}
\label{fig:probe_threshold_selection}
\end{figure}

\paragraph{Probe timestep and consistency threshold.}
Figure~\ref{fig:probe_threshold_selection}(a) shows similar DINO scores across
probe timesteps over a broad PR range, with degradation only under aggressive
compression. This supports an intermediate probe without requiring precise
timestep tuning. In panel (b), subject and background consistency remain
nearly saturated while CMBench DINO responds to lost contextual information,
showing why general video-quality metrics alone cannot select a
recall-preserving compression budget.

\begin{table*}[t]
\centering
\begin{minipage}[t]{0.49\textwidth}
\vspace{0pt}
\centering
\captionof{table}{RoPE re-indexing ablation. $\Delta$DINO is the change from re-indexing.}
\label{tab:reindex_ablation}
\end{minipage}\hfill
\begin{minipage}[t]{0.49\textwidth}
\vspace{0pt}
\centering
\captionof{table}{Token-selection ablation. Reverse retains low-discrepancy tokens.}
\label{tab:selection_direction_ablation}
\end{minipage}
\par\vspace{6pt}\noindent
\begin{minipage}[t]{0.49\textwidth}
\vspace{0pt}
\centering
{\small
\setlength{\tabcolsep}{4.0pt}
\renewcommand{\arraystretch}{1.14}
\resizebox{\linewidth}{!}{%
\begin{tabular}{@{}lrrr@{}}
\toprule
Method &
\shortstack{w/ re-index\\DINO $\uparrow$} &
\shortstack{w/o re-index\\DINO $\uparrow$} &
$\Delta$DINO \\
\midrule
FullKV & 0.6804 & 0.4385 & +0.2418 \\
DeCoPrune--HS (ours) & 0.7156 & 0.6686 & +0.0470 \\
TempDiff & 0.6365 & 0.6019 & +0.0346 \\
Random & 0.6723 & 0.5515 & +0.1208 \\
\midrule
DummyForcing & 0.4005 & 0.4012 & $-0.0007$ \\
ForcingKV & 0.4825 & 0.4860 & $-0.0035$ \\
Streaming & 0.3994 & 0.4040 & $-0.0046$ \\
\bottomrule
\end{tabular}%
}
}
\end{minipage}
\hfill
\begin{minipage}[t]{0.49\textwidth}
\vspace{0pt}
\centering
{\small
\setlength{\tabcolsep}{5.0pt}
\renewcommand{\arraystretch}{1.14}
\resizebox{\linewidth}{!}{%
\begin{tabular}{@{}lrr@{}}
\toprule
Policy &
DINO $\uparrow$ &
PR $\uparrow$ \\
\midrule
DeCoPrune & 0.6701 & 85.43\% \\
Random & 0.6091 & 85.31\% \\
Reverse criterion & 0.5621 & 18.79\% \\
Reverse criterion (matched PR) & 0.4512 & 85.39\% \\
\bottomrule
\end{tabular}%
}
}
\end{minipage}
\end{table*}

\paragraph{Effect of RoPE re-indexing.}
Within the diagnostic evaluation of Table~\ref{tab:reindex_ablation}, we
isolate the effect of mapping the temporal
coordinates of retained historical keys into a compact virtual span. Re-indexing
raises DINO for the methods that preserve content from distant history in this
evaluation, including FullKV, DeCoPrune--HS, TempDiff, and Random. The largest
gain occurs for FullKV, whose uncompressed history contains the greatest
query--key positional offsets. In contrast, Streaming and the two head-wise
policies show little change, with small decreases in DINO. Their cache rules
already discard most distant context. These results indicate that re-indexing improves access
to retained evidence, while the pruning policy determines which evidence
remains available.

\paragraph{Token-selection criterion.}
Table~\ref{tab:selection_direction_ablation} compares our consistency-based
selection with random pruning and the reverse criterion, which retains tokens
with low step-to-final discrepancy. For the matched-PR reverse baseline, we
independently adjust the reverse-selection threshold to match DeCoPrune's
average pruning ratio. At nearly matched pruning ratios
(85.31--85.43\%), DeCoPrune achieves 0.6701 DINO, compared with 0.6091 for
Random and 0.4512 for reverse selection. The 0.2189 DINO gap to the reverse
criterion isolates the importance of selection direction at a comparable
cache budget. Even when reverse selection retains 81.21\% of the historical
KV entries (PR = 18.79\%), it reaches only 0.5621 DINO. Together, these results support retaining
high-discrepancy tokens rather than low-discrepancy tokens.

%% file: sections/conclusion.tex
\section{Conclusion}

Long-context autoregressive video diffusion requires a growing KV cache,
incurring increasing memory and attention costs. \textsc{DeCoPrune} addresses
this with training-free pruning based on denoising consistency, a
model-intrinsic proxy for token importance. We introduce \textsc{CMBench} to
evaluate long-context information retention, and experiments show that
\textsc{DeCoPrune} improves efficiency while preserving continuation
consistency.

%% file: sections/appendix.tex
\section{RoPE Re-indexing Details and Analysis}
\label{app:rope_reindexing}

Let $q_0$ be the temporal position of the first generated latent frame, equal
to the number of context latent frames, and let $t\in\{0,\ldots,q_0-1\}$ be a
context frame's original position. We use a maximum virtual offset
$L_{\mathrm{virt}}$ and preserve the newest $N_r$ context frames verbatim. For
$q_0>N_r$, define the real offset $\delta_t=q_0-t$ and map
\[
\widetilde t(t)=
\begin{cases}
t, & \delta_t\leq N_r,\\[3pt]
q_0-\left[N_r+(\delta_t-N_r)
\dfrac{L_{\mathrm{virt}}-N_r}{q_0-N_r}\right],
& \delta_t>N_r.
\end{cases}
\]
If $q_0\leq N_r$, the mapping is the identity. Thus, the oldest context frame
is placed at $q_0-L_{\mathrm{virt}}$, the recent tail remains at its true
positions, and older frames are compressed linearly between these endpoints.
The mapping is a function of the original frame position rather than the rank
of a retained token. Consequently, tokens from the same source frame receive
the same virtual position even when different head groups retain different
physical token subsets.

\begin{figure}[htbp]
\centering
\includegraphics[width=\linewidth]{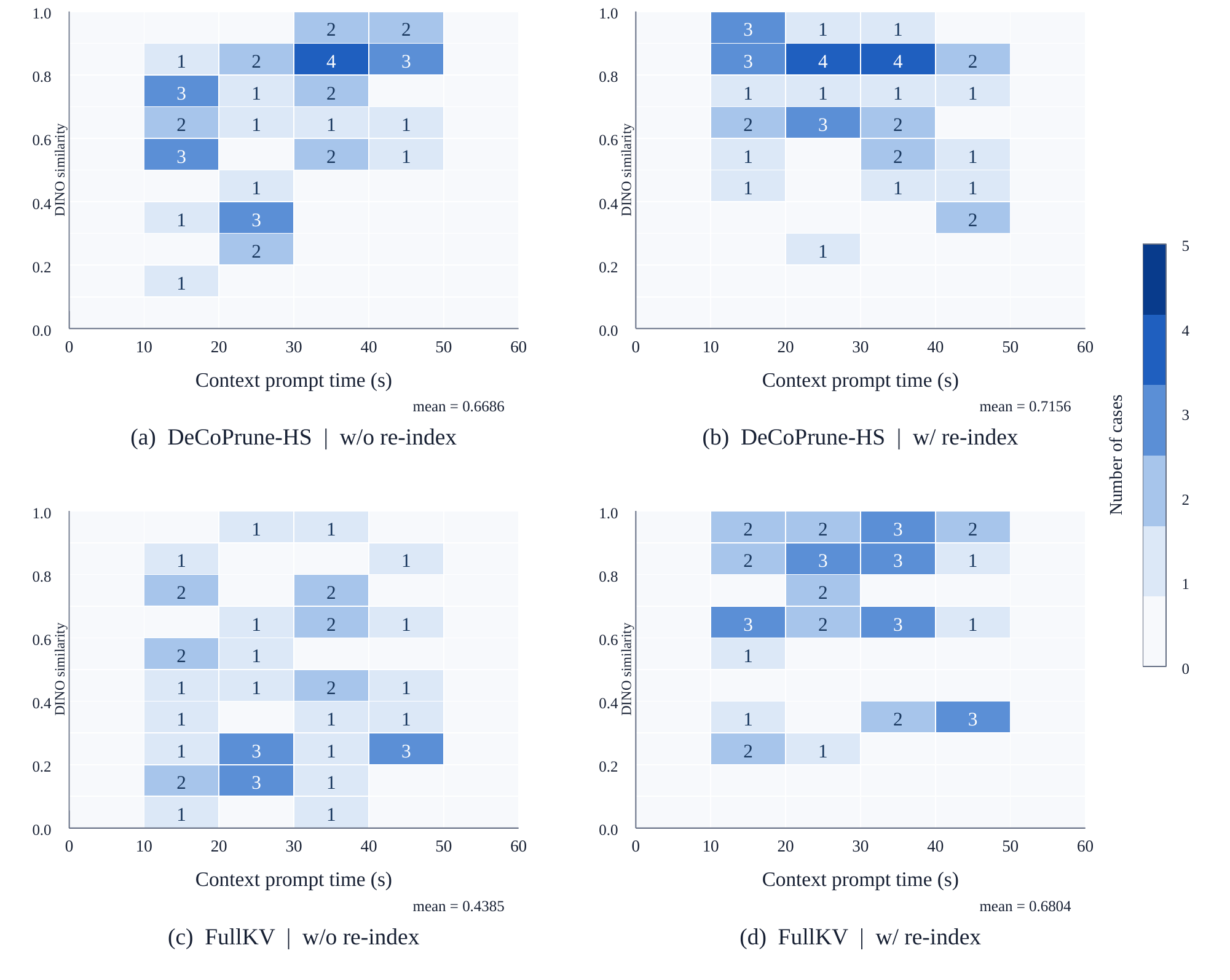}
\caption{\textbf{Effect of RoPE re-indexing across context positions.}
The heatmaps show the distribution of \textsc{CMBench} cases over context-prompt time and DINO similarity for \textsc{DeCoPrune--HS} (top) and \textit{FullKV} (bottom), without re-indexing (left) and with re-indexing (right). Cell values and color intensity indicate the number of cases. In the separate diagnostic evaluation of Table~\ref{tab:reindex_ablation}, re-indexing shifts the score distribution upward for both methods, increasing the mean DINO similarity from 0.6686 to 0.7156 for \textsc{DeCoPrune--HS} and from 0.4385 to 0.6804 for \textit{FullKV}.}
\label{fig:reindex_heatmaps}
\end{figure}

The cached keys already contain RoPE. For temporal frequency pair $f$ with
angular frequency $\theta_f$, we therefore apply the phase correction
\[
\widetilde{\mathbf K}_{t,f}
=\exp\!\left(i\theta_f[\widetilde t(t)-t]\right)\mathbf K_{t,f},
\]
which is equivalent to
$R_{\mathrm{temp}}(\widetilde t)R_{\mathrm{temp}}(t)^{-1}$ on the temporal
RoPE subspace. In our experiments, $L_{\mathrm{virt}}=17$ latent frames,
$N_r=8$, and all temporal frequency pairs are corrected; spatial RoPE pairs
are unchanged. Re-indexing is applied once at the context--generation boundary
after cache compaction. Generation keys are subsequently written at their true
positions. Values, physical token order, and pruning masks are never changed.
Figure~\ref{fig:reindex_heatmaps} visualizes how this correction shifts the
score distribution across context positions for \textsc{DeCoPrune--HS} and
\textit{FullKV}.

\section{Head-Specialized Variant}
\label{app:head_specialization}

\paragraph{Recent local window.}
For the main experiments, the sink and recent windows each contain one chunk
($w=1$).

\textsc{DeCoPrune--HS} combines our token-selection criterion with the
static/dynamic head partition of \textit{ForcingKV}~\citep{forcingkv2026}.
We construct one offline layer--head map from 15 calibration videos using
FullKV attention measured at the first continuation denoising call. For layer
$\ell$ and head $h$, let $A^{(c)}_{\ell,h}(q,k)$ denote softmax attention from
query token $q$ to context key token $k$ in calibration case $c$. Let
$\mathcal{S}_c$ and $\mathcal{R}_c$ contain, respectively, the first and last
four context latent frames. We compute the recent-attention ratio
\[
s_{\ell,h}=
\frac{\sum_c\sum_q\sum_{k\in\mathcal{R}_c}
A^{(c)}_{\ell,h}(q,k)}
{\sum_c\sum_q\sum_{k\notin\mathcal{S}_c}
A^{(c)}_{\ell,h}(q,k)}.
\]
A head is classified as static when $s_{\ell,h}\geq0.8$ and dynamic otherwise.
The resulting partition is fixed and reused throughout evaluation; it is not
recomputed from the test continuation.

For layers 1--39, dynamic heads use the same chunk-level \textsc{DeCoPrune}
mask as the standard variant, with the denoising-consistency threshold
$\gamma=0.10$. Static heads use a streaming cache containing the first four
and most recent four context latent frames. Layer 0 is treated uniformly: all
40 heads follow \textsc{DeCoPrune}, rather than the offline split. After this
layer-0 override, the 40-layer, 40-head model contains 232 static and 1368
dynamic layer--heads. The two groups are packed into separate physical KV
banks, while generated KV entries are appended under the same continuation
policy. The reported PR therefore counts retained tokens separately for every
layer and head, as defined in Equation~\ref{eq:cmbench_pruning_ratio}.

\section{Backbone Selection for \textsc{CMBench}}
\label{app:backbone_selection}

\begin{figure}[!t]
\centering
\includegraphics[width=\linewidth]{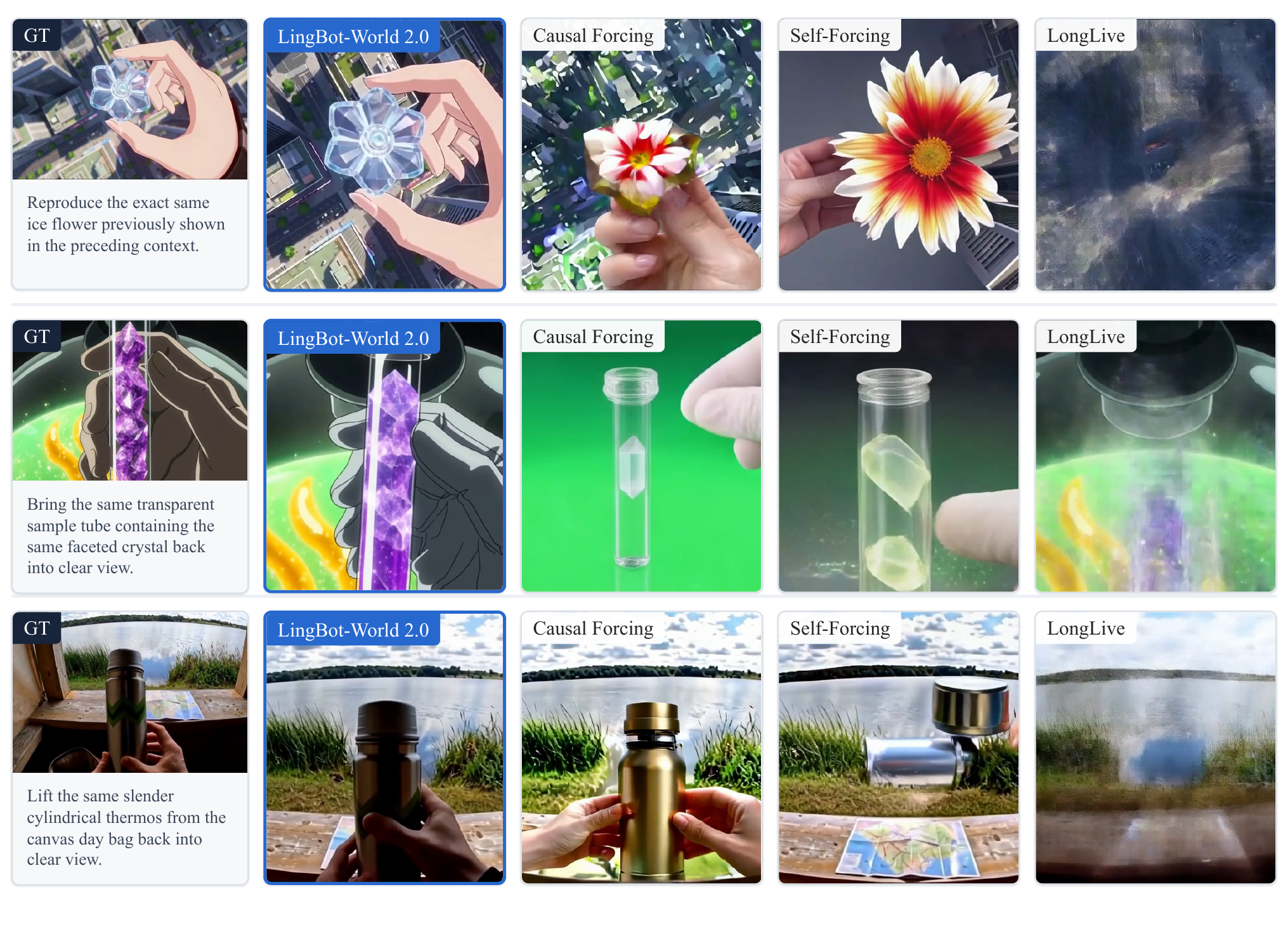}
\caption{\textbf{Full-context comparison across autoregressive video backbones.}
We compare ground-truth reference frames with FullKV generations from
\textit{LingBot World v2}, \textit{Causal Forcing}, \textit{Self-Forcing}, and
\textit{LongLive} on three representative object-retrieval cases that require
no camera-control input. Despite retaining the complete context, the three
alternative backbones frequently fail to recover the target object, whereas
the \textit{LingBot World v2} generations more closely reproduce the targets
in these examples.}
\label{fig:appendix_backbone_qualitative}
\end{figure}

We initially considered \textit{LongLive}~\citep{longlive2025},
\textit{Self-Forcing}~\citep{selfforcing2025}, and
\textit{Causal Forcing}~\citep{causalforcing2026} as backbones for the main
experiments. A suitable backbone for \textsc{CMBench} must remain stable over a
minute-scale rollout and must be able to use information beyond its native
attention span. Otherwise, errors caused by the backbone's own long-horizon
degradation cannot be separated reliably from errors introduced by KV-cache
compression.
Figure~\ref{fig:appendix_backbone_qualitative} shows representative
full-context failures of the alternative backbones, while
Table~\ref{tab:appendix_backbone_selection} reports their quantitative
comparison on the same completed subset.

\makeatletter
\setlength{\@fptop}{0pt}
\makeatother

\begin{table}[htbp]
\centering
\caption{\textbf{Full-context performance of candidate backbones on
\textsc{CMBench}.} Mean DINO (0--1) is computed over the same completed subset for all
backbones.}
\label{tab:appendix_backbone_selection}
\setlength{\tabcolsep}{18pt}
\renewcommand{\arraystretch}{1.12}
\begin{tabular}{@{}lc@{}}
\toprule
\textbf{Backbone} & \textbf{Mean DINO} $\uparrow$ \\
\midrule
LingBot World v2~\citep{lingbotworld2_2026} & \textbf{0.6803} \\
Causal Forcing~\citep{causalforcing2026} & 0.2627 \\
Self-Forcing~\citep{selfforcing2025} & 0.2609 \\
LongLive~\citep{longlive2025} & 0.2608 \\
\bottomrule
\end{tabular}
\end{table}

Table~\ref{tab:appendix_backbone_selection} shows that the three alternative
backbones obtain FullKV DINO scores between 0.2608 and 0.2627 on the completed
subset. This would confound a controlled pruning
study because a missing target could reflect either cache compression or the
generator's inability to use the available context. We therefore use
\textit{LingBot World v2}~\citep{lingbotworld2_2026} as the main backbone.

\subsection{LongLive Continuation with 10-Second Contexts}
\label{app:longlive_short_context}

We additionally evaluate continuation generation with
\textit{LongLive}~\citep{longlive2025} on 18 cases, each conditioned on a
10-second video context. Table~\ref{tab:longlive_short_context} reports the
PR and DINO scores. This shorter-context
experiment is distinct from the minute-scale backbone comparison above.
With $\gamma=0.10$, \textsc{DeCoPrune} achieves 0.5880 DINO at 57.25\%
PR, outperforming the other pruning baselines in both metrics.

\begin{table}[htbp]
\centering
\caption{\textbf{LongLive continuation with 10-second contexts on 18 cases.}
DINO is reported on a 0--1 scale.}
\label{tab:longlive_short_context}
\small
\setlength{\tabcolsep}{18pt}
\renewcommand{\arraystretch}{1.18}
\begin{tabular}{@{}lrr@{}}
\toprule
\textbf{Method} & \textbf{PR} $\uparrow$ & \textbf{DINO} $\uparrow$ \\
\midrule
FullKV & 0.00\% & 0.6106 \\
\textsc{DeCoPrune} & 57.25\% & 0.5880 \\
Streaming & 53.57\% & 0.5162 \\
ForcingKV & 54.17\% & 0.5302 \\
DummyForcing & 55.60\% & 0.5157 \\
\bottomrule
\end{tabular}
\end{table}

\section{Synthetic and Real Video Contexts}
\label{app:video_source}

H3-generated contexts allow explicit control over event timing, target
visibility, and camera transitions, making it possible to isolate retrieval
of information absent from the recent context. The reference is the target
actually visible in the context video, so evaluation measures consistency
with observed evidence rather than agreement with the synthesis prompt.
Real video contexts provide a complementary check that the comparison does
not depend on the visual characteristics of H3 outputs.

Table~\ref{tab:video_source} compares mean DINO scores on the synthetic-video
and real-video subsets under the same evaluation protocol. This
source-stratified analysis uses a single random seed, whereas the main results
in Table~\ref{tab:cmbench_results} are averaged over three seeds. Its values
should therefore be compared within this table rather than pooled to
reconstruct the three-seed main result. Real videos yield
lower scores across all methods, indicating greater difficulty, while
\textsc{DeCoPrune} achieves higher DINO than the other pruning methods on both
subsets. This agreement supports the use of H3-generated
contexts alongside more challenging real videos.

\begin{table}[htbp]
\centering
\caption{\textbf{Evaluation by context-video source.} Mean DINO scores
(0--1) on synthetic-video and real-video contexts using one random seed.}
\label{tab:video_source}
\small
\setlength{\tabcolsep}{12pt}
\begin{tabular}{@{}lcc@{}}
\toprule
\textbf{Method} & \textbf{Synthetic Video} & \textbf{Real Video} \\
\midrule
FullKV & 0.6873 & 0.6621 \\
\textsc{DeCoPrune} & 0.6909 & 0.6412 \\
TempDiff~\citep{hwang2024everest,fu2025framefusion} & 0.6484 & 0.5212 \\
ForcingKV~\citep{forcingkv2026} & 0.5422 & 0.3839 \\
Streaming~\citep{streamingvlm2026,longlive2025} & 0.4592 & 0.3274 \\
DummyForcing~\citep{dummyforcing2026} & 0.4549 & 0.3111 \\
\bottomrule
\end{tabular}
\end{table}

\section{Pruning-Ratio Aggregation}
\label{app:pr_aggregation}

In Equation~\ref{eq:cmbench_pruning_ratio}, the history for generated chunk
$i$ comprises the observed context and all previously completed generated
chunks. The current noisy chunk is excluded from both the compressed and
FullKV counts. Each retained token is counted once per layer and head in
which it is active, accounting for head-specific pruning.

Equivalently, the sequence-level PR is the FullKV-workload-weighted average
of the per-chunk, per-layer, per-head pruning ratios, with weights
proportional to $k_{i,\ell,h}^{\mathrm{full}}$. Thus, attention calls with
longer histories contribute proportionally more to the cumulative count.
We compute this ratio separately for each continuation and then take its
arithmetic mean over evaluation cases; we do not pool token counts across
cases before taking the ratio.

\section{Limitations}

The current study has three main limitations. First, although
\textsc{DeCoPrune} reduces cumulative historical KV token counts by 85.43\% in the main
setting, the retained cache
still grows with autoregressive generation length; memory use and attention
cost are therefore not strictly bounded. Second, few open-source baselines
currently support reliable long-context video generation. Even a comparatively
capable model such as \textit{LingBot World v2} has a measurable intrinsic ceiling on
context consistency, which makes pruning-induced degradation difficult to
fully separate from backbone generation errors. Third, denoising consistency
is a model-intrinsic, future-agnostic signal: it does not explicitly encode the
relevance of a token to a future prompt or task and may therefore undervalue a
rare detail that is easy to denoise now but requested later. These limitations
motivate bounded-memory extensions and combinations with query- or task-aware
memory selection.